\documentclass[pdflatex,sn-mathphys-num,iicol]{sn-jnl}

\usepackage{graphicx}%
\usepackage{multirow}%
\usepackage{amsmath,amssymb,amsfonts}%
\usepackage{amsthm}%
\usepackage[title]{appendix}%
\usepackage{xcolor}%
\usepackage{textcomp}%
\usepackage{manyfoot}%
\usepackage{booktabs}%
\usepackage{algorithm}%
\usepackage{algorithmicx}%
\usepackage{algpseudocode}%
\usepackage{listings}%
\usepackage{comment}

\theoremstyle{thmstyleone}%
\newtheorem{theorem}{Theorem}
\newtheorem{proposition}[theorem]{Proposition}%

\theoremstyle{thmstyletwo}%
\newtheorem{example}{Example}%
\newtheorem{remark}{Remark}%

\theoremstyle{thmstylethree}%
\newtheorem{definition}{Definition}%

\begin{document}

\title[From Performance to Practice: Knowledge-Distilled Segmentator for On‑Premises Clinical Workflows]{From Performance to Practice: Knowledge-Distilled Segmentator for On‑Premises Clinical Workflows}


\author[1]{\fnm{Qizhen} \sur{Lan}}
\author[2]{\fnm{Aaron} \sur{Choi}}
\author[3]{\fnm{Jun} \sur{Ma}}
\author[3]{\fnm{Bo} \sur{Wang}}
\author[4]{\fnm{Zhongming} \sur{Zhao}}
\author[1]{\fnm{Xiaoqian} \sur{Jiang}}
\author*[1]{\fnm{Yu-Chun} \sur{Hsu}}\email{yu-chun.hsu@uth.tmc.edu}

\affil[1]{\orgname{D. Bradley McWilliams School of Biomedical Informatics},
\orgname{The University of Texas Health Science Center at Houston},
\city{Houston}, \state{TX}, \postcode{77030}, \country{USA}}

\affil[2]{\orgname{M31 AI},
\city{Toronto}, \country{Canada}}

\affil[3]{\orgname{University Health Network},
\orgname{University of Toronto},
\orgname{Vector Institute},
\city{Toronto}, \country{Canada}}

\affil[4]{\orgname{Center for Precision Health},
\orgname{D. Bradley McWilliams School of Biomedical Informatics},
\orgname{The University of Texas Health Science Center at Houston},
\city{Houston}, \state{TX}, \postcode{77030}, \country{USA}}


\abstract{Deploying medical image segmentation models in routine clinical workflows is often constrained by on-premises infrastructure, where computational resources are fixed and cloud-based inference may be restricted by governance and security policies. While high-capacity models achieve strong segmentation accuracy, their computational demands hinder practical deployment and long-term maintainability in hospital environments.
We present a deployment-oriented framework that leverages knowledge distillation to translate a high-performing segmentation model into a scalable family of compact student models, without modifying the inference pipeline. The proposed approach preserves architectural compatibility with existing clinical systems while enabling systematic capacity reduction.
The framework is evaluated on a multi-site brain MRI dataset comprising 1,104 3D volumes, with independent testing on 101 curated cases, and is further examined on abdominal CT to assess cross-modality generalizability. Under aggressive parameter reduction (94\%), the distilled student model preserves nearly all of the teacher’s segmentation accuracy (98.7\%), while achieving substantial efficiency gains, including up to a 67\% reduction in CPU inference latency without additional deployment overhead.
These results demonstrate that knowledge distillation provides a practical and reliable pathway for converting research-grade segmentation models into maintainable, deployment-ready components for on-premises clinical workflows in real-world health systems.
}

\maketitle

\section{Introduction}\label{sec:intro}

Deep learning (DL) has become a core methodology in medical imaging research and is increasingly evaluated for use in routine clinical workflows, with strong performance reported across classification, segmentation, risk stratification, and related tasks \cite{aggarwal2021diagnostic, mall2023comprehensive, li2023medical}. Despite these advances, high-capacity models remain far more prevalent in experimental studies than in routine clinical use, indicating that improvements in accuracy alone have not translated into widespread adoption in real-world healthcare settings. In these settings, limitations in infrastructure and operational constraints often govern deployment decisions, outweighing marginal gains in algorithmic accuracy when determining whether AI tools can be sustainably deployed.
\begin{figure*}[htb]
    \centering
    \includegraphics[width=0.95\linewidth]{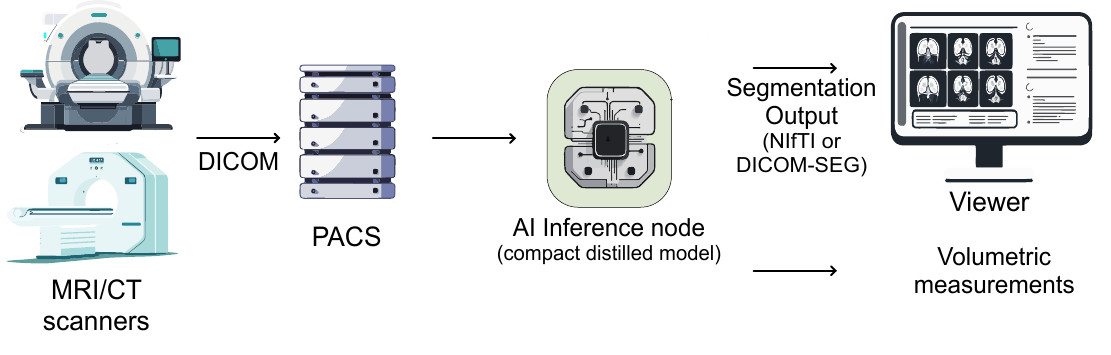}
    \caption{
    Deployment pathway of the compact distilled segmentation model within an on-premises clinical environment. Imaging data are retrieved from PACS and processed on a local inference node running the distilled model. Segmentation outputs are returned in standard clinical formats (e.g., NIfTI or DICOM-SEG) for visualization and downstream quantitative analysis. This figure illustrates a representative integration scenario and does not correspond to a specific experimental pipeline.
    }
    \label{fig:deployment}
\end{figure*} 
Recent studies suggest that barriers to clinical adoption are increasingly due to challenges in deployment and integration rather than to limitations in model precision \cite{bajwa2021artificial, lekadir2025future}. The deployment of AI systems in health systems requires interoperability with existing hospital infrastructure, compliance with governance and privacy regulations, and stable behavior between scanners and acquisition protocols under operational constraints that differ substantially from research environments. A review of 140 radiology studies reported that only a small fraction evaluated real-world implementation or workflow integration \cite{lawrence2025artificial}. Similarly, a European survey found that although radiologists are encountering AI tools more frequently, familiarity with regulatory obligations and post-market requirements remains limited \cite{zanardo2024impact}. These findings highlight a persistent gap between the algorithmic performance reported and the practical requirements of clinical implementation.

This gap is further amplified in hospital environments on-premises. Many healthcare systems operate behind strict firewalls and restrict cloud-based computation due to privacy and security policies, necessitating local inference on fixed hardware resources. In practice, inference is often performed on CPU-dominant systems or a single consumer-grade GPU, with limited memory and strict latency constraints. Under these conditions, computational efficiency is not merely an optimization objective but a prerequisite for safe, sustainable deployment within health-system infrastructure. While several PACS-integrated AI systems have demonstrated that workflow integration is feasible when infrastructure compatibility and reliability are prioritized \cite{zhang2023vendor, sandhu2020integrating, patterson2019scope}, these solutions are typically task-specific and do not provide a generalizable pathway for adapting high-capacity models to resource-constrained clinical environments.

Model compression offers a potential strategy to address these system-level constraints. Approaches such as structured pruning \cite{yang2025pruning, valverde2024sauron}, quantization \cite{askarihemmat2019u}, and architectural simplification \cite{lu2022half} can substantially reduce model size and computational cost. However, these methods often require task-specific tuning or structural modification, which can complicate validation, maintenance, and regulatory compliance over the model lifecycle in clinical settings. In contrast, knowledge distillation (KD) provides a model-agnostic mechanism to transfer predictive behavior from a large teacher network to a compact student model \cite{mansourian2025comprehensive, lan2024gradient, lan2025acam}. In medical imaging, KD has been applied to improve lightweight segmentation networks \cite{zhao2023mskd, lan2026recokdregioncontextawareknowledge}, maintain performance during continual updates \cite{gonzalez2023lifelong}, and enhance robustness in heterogeneous acquisition settings.

From a deployment perspective, logit-based knowledge distillation is particularly attractive for translational systems. It introduces no architectural constraints, does not require alignment at the feature-level, and imposes minimal engineering overhead \cite{hsu2022closer, hinton2015distilling}. These properties simplify training and validation, improve reproducibility, and reduce system complexity. Such characteristics are especially important in regulated healthcare settings where model updates and maintenance must be predictable and auditable.

\begin{figure*}[h]
    \centering
    \includegraphics[width=0.98\linewidth]{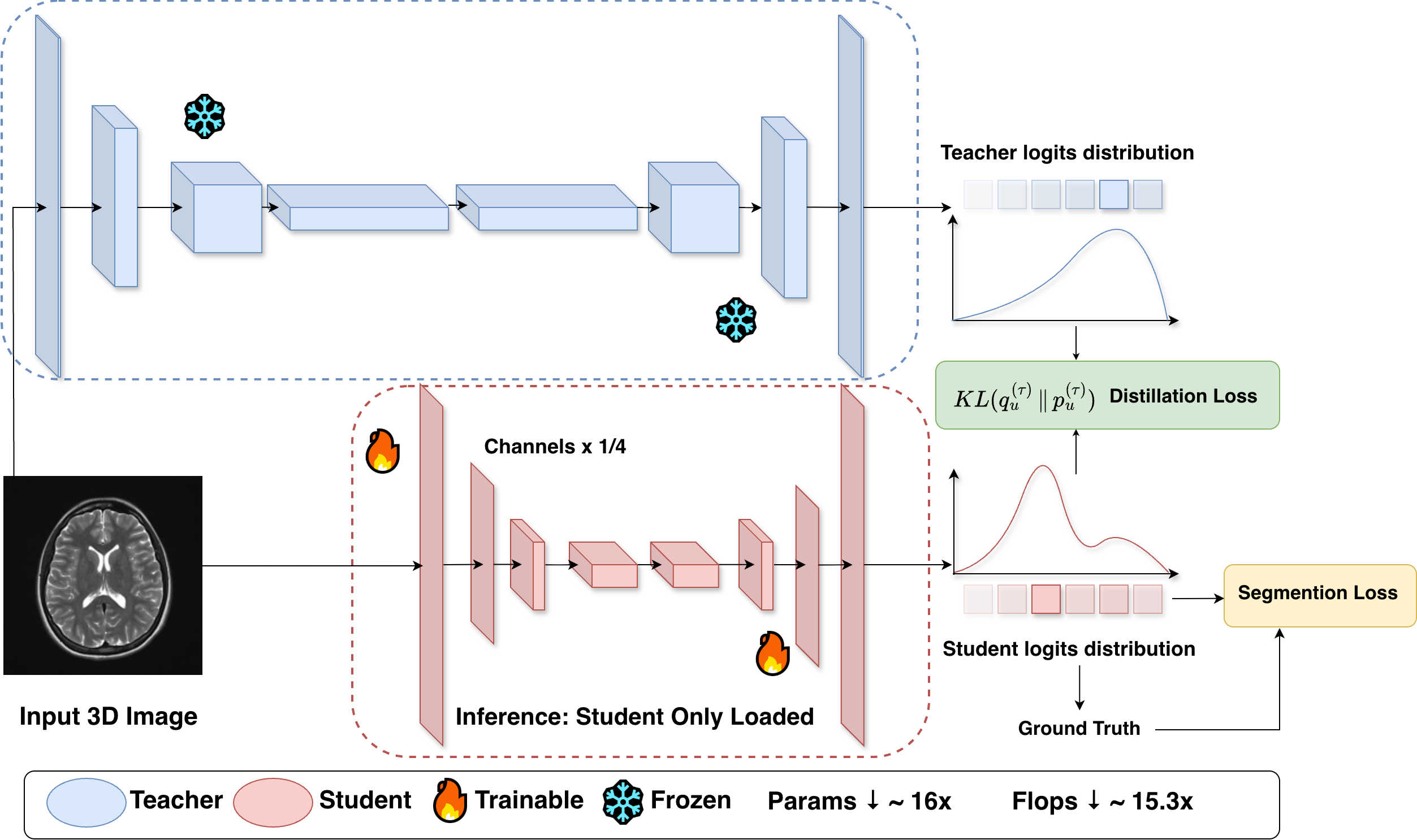}
    \caption{
    Overview of the proposed logit-based knowledge distillation framework for deployment-oriented model compression. A high-capacity nnU-Net teacher produces softened outputs to supervise a compact student via a KL-divergence loss, in addition to standard segmentation supervision. The teacher is used only during training, while the distilled student model is deployed for on-premises inference, enabling efficiency gains without altering the inference pipeline.
    }
    \label{fig:kd-overview}
\end{figure*}

In this study, we investigate how logit-based knowledge distillation can be used to translate a high-performing nnU-Net teacher into compact, deployment-ready student models for on-premises clinical workflows. Figure~\ref{fig:deployment} provides an overview of the proposed deployment pathway, illustrating how the compact distilled model integrates with standard PACS-based systems for clinical inference and downstream analysis. Figure~\ref{fig:kd-overview} illustrates the proposed distillation-based framework, in which a high-capacity nnU-Net teacher \cite{isensee2021nnu} guides a compact student during training, while only the distilled student is retained for inference. Our objective is not merely to compress a segmentation model, but to establish a practical pathway from research-grade performance to operational feasibility under fixed system constraints. Specifically, we:
\begin{itemize}
    \item establish a unified logit-based KD framework for distilling a high-capacity nnU-Net teacher into compact student models suitable for on-premises deployment;
    \item enable a scalable family of student models through uniform channel reduction while preserving architectural compatibility with existing preprocessing and deployment pipelines;
    \item demonstrate the framework on multi-site brain MRI and assess its generalizability on abdominal CT (BTCV), showing that KD mitigates performance degradation under aggressive compression while substantially improving inference efficiency.
\end{itemize}

Our framework outline a deployment-oriented pathway for converting research-grade medical image segmentation models into reliable and maintainable components of routine on-premises clinical workflows. 

\section{Results}\label{sec:results}
This section reports results with respect to deployment feasibility under on-premises constraints, focusing on whether knowledge distillation (KD) enables stable and efficient operating points for compressed segmentation models. Unless otherwise noted, comparisons between distilled and non-distilled models are performed at the same scale factor to isolate the effect of distillation.

\subsection{Stability of compressed models under on-premises constraints}\label{sec:seg}
\begin{table*}[ht]
\centering
\caption{Segmentation performance on the Mindboggle-101 dataset, evaluated using mean Dice, normalized surface Dice (NSD), and HD95. Results are averaged over 101 human-corrected test cases. Higher Dice and NSD, and lower HD95 indicate better performance.}
\label{tab:brain}
\begin{tabular}{lccccc}
\toprule
Model & Scale & Params (M) & Dice $\uparrow$ & NSD $\uparrow$ & HD95 $\downarrow$ \\
\midrule
Teacher & -- & 102.44 & 81.65 & 90.48 & 2.82 \\

Student & $\times$1/2 & 25.64 & 79.48 & 90.12 & 2.92 \\
Student + KD   & $\times$1/2 & 25.64 & 81.30 & 90.29 & 2.87 \\
Student & $\times$1/4 & 6.43 & 78.92 & 89.66 & 3.18 \\
Student + KD   & $\times$1/4 & 6.43 & 80.58 & 89.85 & 2.96 \\
\bottomrule
\end{tabular}
\end{table*}

\textbf{Take away 1:} \textit{Knowledge distillation stabilizes aggressively compressed segmentation models, preserving clinically meaningful fidelity at operating points that would otherwise be unreliable for deployment.} 

Table~\ref{tab:brain} summarizes segmentation performance on the independent Mindboggle-101 benchmark under systematic channel reduction. As model capacity is reduced, non-distilled student models exhibit predictable degradation in both overlap- and boundary-based metrics, indicating brittle behavior under compression. In contrast, distilled students retain a high fraction of teacher performance across compression levels. At $\times$1/2 scale, KD preserves nearly all segmentation fidelity (99.6\%) relative to the teacher, while under more aggressive compression ($\times$1/4), KD substantially mitigates the loss induced by capacity reduction (recovering 60.8\% of the lost performance).

Boundary-sensitive metrics reveal a clearer distinction between distilled and non-distilled models. While normalized surface Dice (NSD) changes remain modest, KD markedly reduces degradation in HD95 as compression increases, indicating improved boundary stability rather than uniform inflation of overlap scores. This distinction is particularly relevant in clinical contexts, where boundary inconsistencies and local structural errors can undermine trust in automated outputs.

From a system perspective, these results indicate that KD transforms compressed models from brittle configurations into stable operating points. This stabilization does not depend on modifying the inference pipeline or introducing additional runtime components, making aggressive capacity reduction compatible with the reliability requirements of on-premises clinical workflows.

\subsection{Computational efficiency under on-premises constraints}
\begin{table*}[ht]
\centering
\caption{Efficiency metrics (lower is better). CPU time averaged on MindBoggle-101 using Intel Xeon Silver 4116; GPU on 2080 Ti and H100. Knowledge distillation affects training only; distilled and non-distilled students at the same scale have identical inference-time cost.}
\label{tab:efficiency}
\setlength{\tabcolsep}{4.5pt}
\begin{tabular}{lccccccc}
\toprule
Model & Scale & Params (M) & GFLOPs & Mem (G) & CPU (s) & 2080 Ti & H100 \\
\midrule
Teacher & -- & 102.44 & 3364.88 & 14.00 & 82.67 & 2.07 & 1.02\\
Student & $\times$1/2 & 25.64 & 853.28 & 8.95 & 45.84  & 1.45 & 0.67\\
Student + KD   & $\times$1/2 & 25.64 & 853.28 & 8.95 & 45.84  & 1.45 & 0.67\\
Student & $\times$1/4 & 6.43  & 219.35 & 6.49 & 27.21  & 1.16 & 0.53\\
Student + KD   & $\times$1/4 & \textbf{6.43} & \textbf{219.35} & \textbf{6.49} &
\textbf{27.21} & \textbf{1.16} & \textbf{0.53}\\
\bottomrule
\end{tabular}
\end{table*}

\textbf{Take away 2:} \textit{Channel reduction yields substantial, hardware-aligned efficiency gains that enable practical on-premises deployment, while knowledge distillation preserves segmentation stability without increasing runtime complexity.}

Table~\ref{tab:efficiency} reports model size, computational cost, memory footprint, and inference latency across representative CPU- and GPU-based platforms. Uniform channel reduction leads to substantial structural savings. Relative to the teacher model, the $\times$1/2 and $\times$1/4 students reduce parameter count and GFLOPs by approximately 75\% and 94\%, respectively, reflecting the quadratic scaling of computation with channel width. Memory footprint is reduced by 36\% at $\times$1/2 and 54\% at $\times$1/4, further improving suitability for resource-constrained environments.

These reductions translate into meaningful runtime improvements, particularly on CPU-dominant systems. On CPU, the $\times$1/2 student achieves an approximately 1.8$\times$ speedup (44.5\% lower latency), while the $\times$1/4 student reaches a 3.0$\times$ speedup (67.1\% lower latency) compared with the teacher. This shift moves inference from a regime that is often impractical for routine clinical use toward operating points that better align with on-premises workflows. GPU measurements on both a consumer 2080~Ti and an H100 show consistent trends, with latency reductions ranging from roughly 30\% to 48\% depending on hardware and scale factor. Because knowledge distillation affects training only, it improves segmentation accuracy without changing inference-time cost or deployment complexity at a fixed scale.

Taken together with the segmentation results in \ref{sec:seg}, these findings delineate a practical operating envelope for on-premises clinical deployment: channel reduction delivers substantial efficiency gains, and KD preserves segmentation fidelity within this constrained computational budget. This combination is particularly relevant for PACS-integrated pipelines, where inference latency, memory usage, and reliability must remain stable across model updates and hardware configurations.

\subsection{Framework-level generality across imaging modalities (CT)}\label{sec:btc}
\begin{table*}[ht]
\centering
\caption{
Organ-wise Dice scores for 13 abdominal structures on the BTCV dataset, along with overall mean Dice (mDice) and HD95.
Results compare the teacher, uniformly compressed student baselines, and their distilled counterparts under identical computational budgets.
}
\resizebox{\textwidth}{!}{
\begin{tabular}{lcccccccccccccccc}
\toprule
\textbf{Model} & \textbf{Scale} &\textbf{Spl} & \textbf{RKid} & \textbf{LKid} & \textbf{Gall} & \textbf{Eso} & \textbf{Liv} & \textbf{Sto} & \textbf{Aor} & \textbf{IVC} & \textbf{Veins} & \textbf{Pan} & \textbf{Rad} & \textbf{Lad} & \textbf{mDice} & \textbf{HD95} \\
\midrule
\textbf{Teacher} & --& 96.39 & 94.72 & 94.92 & 73.84 & 80.32 & 97.33 & 87.16 & 90.68 & 87.82 & 75.21 & 84.51 & 72.94 & 77.40 & 85.64 & 4.66\\
\midrule
\textbf{Student} & $\times 1/2$& 96.02 & 94.18 & 94.68 & 62.80 & 79.96& 96.47& 88.78& 89.79& 86.78& 71.26& 82.08& 68.10& 74.37 & 83.48 & 6.54\\
\textbf{student + KD}  & $\times 1/2$& 96.33 & 94.77 & 94.98 & 61.88 & 83.79 & 96.59 & 84.78 & 91.89 & 88.57 & 72.63 & 83.87 & 68.34 & 76.87 & 84.25 & 6.46\\
\textbf{Student} & $\times 1/4$ & 89.30 & 93.52 & 93.33 & 58.82 & 77.41 & 96.11 & 74.97 & 89.78 & 86.50 & 65.58 & 81.19 & 66.79 & 71.66 & 80.38 & 15.19\\
\textbf{student + KD}  & $\times 1/4$ & 90.17 & 89.67 & 85.54 & 58.45 & 80.30 & 96.10 & 82.96 & 91.68 & 86.41 & 70.76 & 79.92 & 67.53 & 76.02 & 82.79 & 7.59\\
\bottomrule
\end{tabular}
}
\vspace{-1em}
\label{tab:btcv}
\end{table*}

\textbf{Take away 3:} \textit{The proposed distillation-based deployment pipeline generalizes across imaging modalities without task-specific redesign, supporting reuse in multi-service clinical environments.}

To assess framework-level generality beyond the original training anatomy, we applied the same student–teacher configuration to abdominal CT segmentation using the BTCV dataset. As summarized in Table~\ref{tab:btcv}, the observed performance trends closely mirror those seen in brain MRI. At matched computational budgets, distilled student models consistently outperform their non-distilled counterparts, with the performance gap widening under more aggressive capacity reduction. Larger anatomical structures remain relatively robust to compression, whereas smaller or lower-contrast structures exhibit greater sensitivity and benefit more substantially from knowledge distillation. This evaluation is not intended to establish optimal cross-modality performance. Rather, it demonstrates that the proposed compression and distillation pipeline can be transferred to a new imaging modality and anatomical domain without modifying the model architecture, distillation objectives, or deployment configuration. This invariance supports the use of knowledge-distillation–based compression as a reusable, modality-agnostic component within clinical systems that support multiple imaging services.




\subsection{Qualitative failure modes relevant to clinical workflows}
\begin{figure}[ht]
    \centering
    \includegraphics[width=0.99\linewidth]{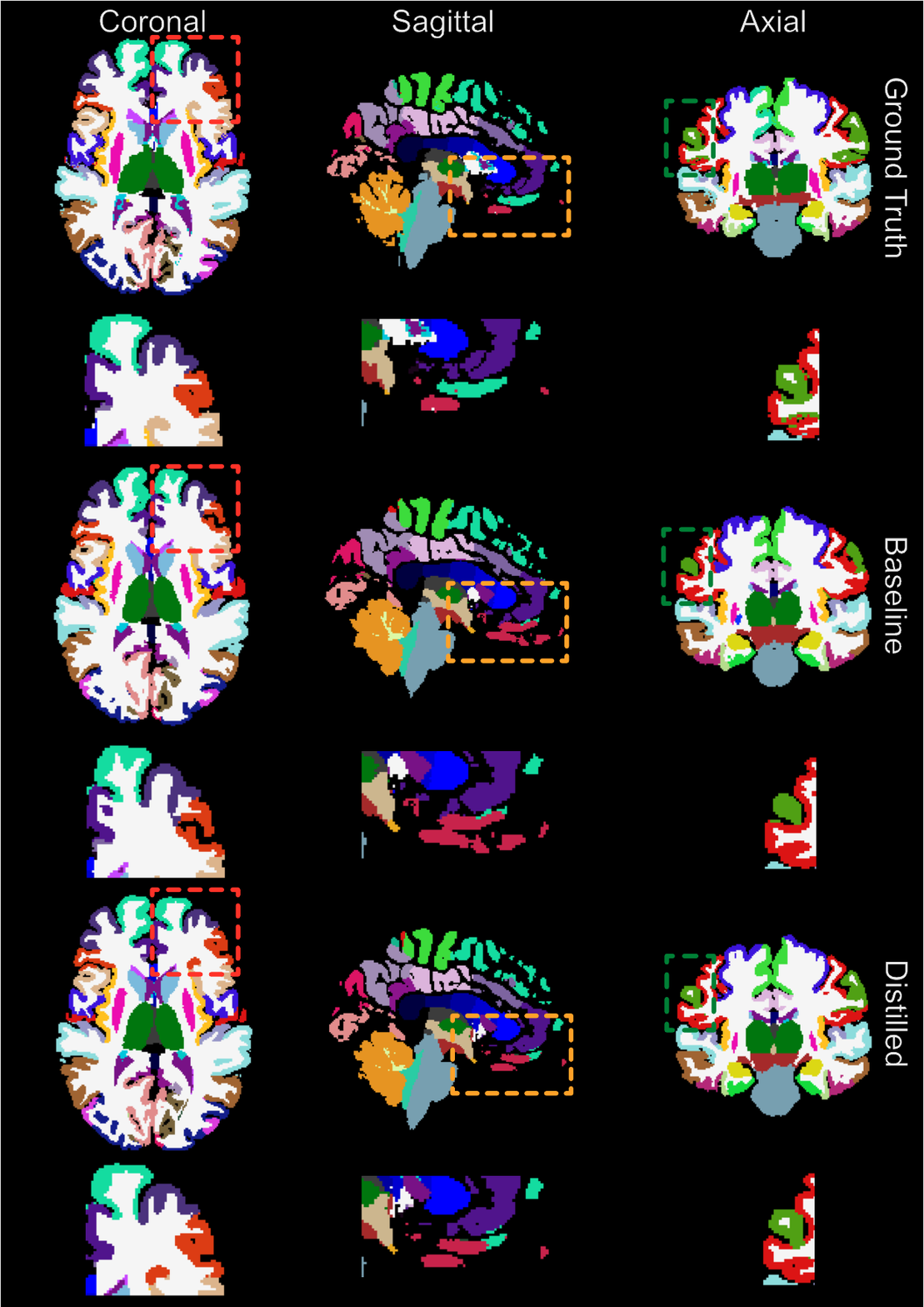}
    \caption{
    Representative qualitative comparisons across coronal, sagittal, and axial views.
    Ground truth annotations (top), non-distilled student predictions (middle), and distilled student predictions (bottom) are shown. Dashed regions highlight areas where differences between methods are visually apparent. The distilled model exhibits improved local consistency and boundary preservation relative to the non-distilled baseline, consistent with the quantitative trends.
    }
    \label{fig:qualitative}
\end{figure}

\textbf{Take away 4:} \textit{Qualitative analysis shows that knowledge distillation reduces clinically relevant failure modes under aggressive compression, improving structural consistency and boundary reliability in routine workflows.}

Figure~\ref{fig:qualitative} presents representative qualitative comparisons across coronal, sagittal, and axial views, contrasting ground-truth annotations with predictions from non-distilled and knowledge-distilled (KD) student models under aggressive capacity reduction. These visualizations complement the quantitative evaluation by revealing characteristic error patterns that are not fully reflected in averaged performance metrics.

Non-distilled student models exhibit systematic degradation as capacity is reduced, including partial omission of small anatomical structures, local fragmentation, and inconsistent delineation across adjacent slices. Such failure modes are most pronounced in fine-grained regions, where limited representational capacity amplifies instability and disrupts structural coherence. These errors often manifest as discontinuous boundaries or slice-wise inconsistency, which can undermine confidence in downstream clinical use.

In contrast, distilled models produce more coherent and anatomically consistent segmentations across views. Knowledge distillation improves the preservation of small structures and maintains boundary continuity without introducing artificial inflation of segmented regions. These qualitative improvements align with boundary-sensitive quantitative trends and indicate that KD stabilizes model behavior in regimes where non-distilled compression becomes brittle.

From a health-system perspective, mitigating instability under compression is clinically meaningful. Small structures and boundary accuracy frequently play a disproportionate role in downstream measurements, visualization, and clinical interpretation. Improved robustness in these regions supports trust, usability, and interpretability in routine on-premises workflows, where aggressive capacity reduction may be required to satisfy hardware and latency constraints.

\section{Discussion}\label{sec:discussion}

This study examines knowledge distillation (KD) from a health-system deployment perspective, asking whether it can reliably translate research-grade medical image segmentation models into solutions that are feasible, stable, and maintainable within on-premises clinical environments. Rather than targeting incremental performance gains, our findings indicate that logit-based KD preserves clinically meaningful segmentation fidelity while substantially reducing computational cost. This trade-off defines a practical accuracy–efficiency operating envelope aligned with the reliability, latency, and resource constraints of real-world clinical workflows.

A key finding is that KD stabilizes compact models at capacity levels that would otherwise be unreliable for deployment. Under uniform channel reduction, non-distilled student models exhibit predictable degradation, with boundary-sensitive metrics and qualitative inspection revealing brittle behavior such as fragmented contours, slice-wise inconsistency, and partial omission of small anatomical structures. 

In contrast, distilled students preserve a high fraction of teacher performance under aggressive compression without increasing inference cost, thereby recovering much of the loss observed in non-distilled models and converting otherwise unstable configurations into clinically usable operating points. This behavior is consistent with prior analyses showing that KD preserves representational fidelity more effectively than structure- or precision-based compression alone \cite{movva2022combining,dantas2024comprehensive}. In particular, our work extends these findings by demonstrating their practical relevance in clinically realistic segmentation tasks and deployment settings, rather than in language models or synthetic benchmarks.

From a health-system perspective, boundary stability is a critical determinant of clinical usability. Even modest boundary instability can disproportionately affect downstream measurements, visualization, and clinical trust, despite acceptable global overlap metrics. Consistent with this interpretation, the qualitative results show that KD improves structural coherence and boundary continuity across views, aligning with gains observed in boundary-sensitive metrics. Together, these findings indicate that KD mitigates clinically relevant failure modes under aggressive compression, rather than merely inflating aggregate performance scores.

From an operational standpoint, the efficiency gains observed in this study are directly relevant to health-system deployment. On CPU-dominant systems, inference latency is reduced by nearly half at moderate compression and by more than two-thirds under aggressive compression, shifting inference toward regimes compatible with routine on-premises workflows. Unlike many compression techniques that introduce architectural changes or runtime overhead, KD is applied during training and requires no changes to the deployed inference pipeline. This property simplifies model validation, maintenance, and replacement in PACS-integrated environments, where predictability, reproducibility, and stability are often as critical as raw accuracy. By leaving the deployed inference pipeline unchanged, KD minimizes re-validation requirements in regulated clinical settings, addressing system-level constraints that have been widely recognized as barriers to clinical translation  \cite{sandhu2020integrating,zhang2023vendor}.

Conceptually, the benefits of KD observed here are consistent with its role as a form of semantic transfer. By exposing the student to softened output distributions, KD conveys inter-class relationships and structural cues that are not captured by hard labels alone. This mechanism becomes particularly valuable under aggressive compression, where limited model capacity amplifies local instability and boundary errors. Similar interpretations have been proposed in prior studies of KD for medical image segmentation and multi-task learning \cite{hsu2022closer,zhao2023mskd,hinton2015distilling}. Extending this work, our qualitative analyses provide complementary visual evidence that these representational advantages translate into improved preservation of small structures and local anatomical detail under clinically realistic deployment constraints, rather than being limited to experimental or benchmark settings.

We further show the proposed KD framework generalizes across imaging modalities without task-specific redesign. When applied to abdominal CT segmentation, the same teacher-student configuration yields trends consistent with those observed in brain MRI, with distilled students outperforming non-distilled counterparts at matched computational budgets. This evaluation is not intended to establish state-of-the-art cross-modality performance, but to demonstrate that the same KD framework can be reused across services without architectural or workflow changes, an important consideration for multi-service health systems seeking scalable and maintainable AI infrastructure.

Taken together, these findings support KD as a practical, deployment-oriented strategy for bridging the gap between research-grade segmentation performance and operational feasibility in real-world clinical systems. By enabling aggressive capacity reduction while preserving stability, boundary reliability, and deployment simplicity, KD provides a reusable and repeatable pathway for sustaining clinically usable segmentation performance under fixed on-premises constraints. More broadly, this perspective highlights that in routine care settings, infrastructure compatibility, reliability, and long-term maintainability are often critical determinants of the real-world impact of medical AI systems, alongside algorithmic performance.

\section{Methods}
Our objective is to translate a high-performing but computationally intensive nnU-Net segmentation model into an efficient, workflow-ready tool suitable for routine use in on-premises clinical environments. To achieve this, we employ a logit-based knowledge distillation framework that transfers predictive behavior from a high-capacity teacher to a compact student model while preserving architectural compatibility with existing deployment pipelines. Figure~\\ref{fig:kd-overview} provides an overview of the distillation process used in this study.

\subsection{Data} \label{sec:data}

\subsubsection{Multi-site Brain MRI Training Dataset}

To train a stable teacher model suitable for downstream knowledge distillation, we assembled a large multi-site T1-weighted brain MRI dataset designed to capture the heterogeneity encountered in routine clinical imaging. The dataset spans a wide range of ages, scanner vendors, and neurological conditions, providing a robust supervisory signal for learning generalizable anatomical representations. The aggregated training set comprises 1,104 participants drawn from four publicly available cohorts:

\begin{itemize}
\item \textbf{ABIDE I}~\cite{ABIDEI}: 100 subjects (37 females, 63 males; ages 6.5–39 years), evenly split between individuals with autism spectrum disorder and typically developing controls.

\item \textbf{CoRR}~\cite{CoRR}: 418 healthy subjects (228 females, 190 males; ages 6–62 years) acquired across multiple institutions, emphasizing scanner diversity and reproducibility.

\item \textbf{ADNI}~\cite{ADNI}: 100 subjects (50 females, 50 males; ages 55–96 years), including patients with Alzheimer’s disease and age-matched controls.

\item \textbf{SALD}~\cite{SALD}: 486 healthy adults (304 females, 182 males; ages 19–80 years) scanned using a standardized 3T MPRAGE protocol.
\end{itemize}

All images were preprocessed and segmented using FreeSurfer (v7.4.1) to generate cortical and subcortical labels, which serve as reference annotations for training the teacher model. Due to the substantial computational cost of preprocessing, we intentionally subsampled 100 cases from ABIDE I and ADNI. Rather than maximizing sample count from a single cohort, we prioritized cross-dataset diversity in age range and clinical conditions to train a stable teacher suitable for downstream distillation. As all datasets are fully de-identified and publicly available, no additional institutional review board approval or participant consent was required. Collectively, this dataset provides a realistic representation of clinical heterogeneity, supporting the development of a teacher model that can function as a consistent and reliable reference for student distillation. 

\subsubsection{Evaluation Dataset: Mindboggle-101}

Model performance was evaluated on the independent Mindboggle-101 dataset~\cite{klein2005mindboggle}, which contains 101 subjects with expertly curated anatomical labels. This dataset is widely used as a benchmark for structural brain parcellation and provides high-quality manual segmentations for assessing whether compact student models preserve anatomical fidelity after compression and distillation.

\subsubsection{Generalization Dataset: BTCV Abdominal CT}

To assess whether the proposed distillation framework can be applied beyond the original training modality and anatomy, we further evaluated the student model on the Beyond the Cranial Vault (BTCV) abdominal CT dataset~\cite{wang2012multi}. BTCV includes 30 contrast-enhanced CT volumes with annotations for 13 abdominal organs, following a standard 24/6 train–validation split. Crucially, the student and teacher architectures and the KD framework remain unchanged in this setting. This experiment is not intended to optimize cross-modality segmentation performance, but rather to demonstrate that the proposed distillation pipeline can be applied without modification across different imaging modalities and anatomical targets.

\subsubsection{Data Summary}

Table~\ref{tab:data_summary} summarizes the datasets used for training and evaluation. Together, these datasets enable (1) multi-site robustness in the teacher model, (2) high-quality benchmarking using Mindboggle-101, and (3) cross-modality validation using BTCV to demonstrate the general applicability of the proposed KD framework.

\begin{table*}[!ht]
\caption{Summary of datasets used for training, evaluation, and generalization.}
\centering
\begin{tabular}{lccccc}
\toprule
Dataset & Modality & Condition & Age Range & Subjects & Role\\
\midrule
ABIDE I & MRI & ASD/Control & 6.5–39 & 100 & Training\\
CoRR & MRI & Healthy & 6–62 & 418 & Training\\
SALD & MRI & Healthy & 19–80 & 486 & Training\\
ADNI & MRI & Alzheimer’s/Control & 55–96 & 100 & Training\\
Mindboggle-101 & MRI & Healthy & 19–61 & 101 & Evaluation\\
\midrule
BTCV & CT & Abdominal organs & -- & 30 & Training \& Evaluation\\
\bottomrule
\end{tabular}
\label{tab:data_summary}
\end{table*}

\subsection{Problem Setup}

Let $x \in \mathbb{R}^{|\Omega|}$ denote a 3D medical image defined over a voxel
domain $\Omega$, and let $y \in \{1,\ldots,C\}^{|\Omega|}$ denote the corresponding
voxel-wise segmentation labels, where $C$ is the number of anatomical classes.
The teacher and student segmentation models are denoted by $f_T(\cdot)$ and
$f_S(\cdot)$, respectively.

For an input image $x$, the models output voxel-wise logits
$\mathbf{z}_T = f_T(x)$ and $\mathbf{z}_S = f_S(x)$, where each logit vector
$\mathbf{z}^{(u)} \in \mathbb{R}^C$ represents class scores at voxel $u \in \Omega$.
The student is trained using a supervised segmentation objective together with
a distillation term that encourages agreement with the teacher’s output
distribution.

\subsection{Teacher Model: Multi-site nnU-Net}
To serve as a stable supervisory signal, we train a 3D nnU-Net teacher on the aggregated multi-site brain MRI training set described in Sec.~\ref{sec:data}. The teacher serves as a stable supervisory signal for distillation. nnU-Net provides a strong baseline by adapting key training and architectural choices (e.g., patch size, normalization, depth, and augmentation) to the data distribution, reducing the need for task-specific manual tuning.

Ground-truth cortical and subcortical labels are derived from FreeSurfer processing (v7.4.1) and used as reference annotations for training. After training, the teacher model is frozen and used only to generate voxel-wise soft targets for student distillation.

\subsection{Student Model: Scalable Lightweight Design for Clinical Environments}

\paragraph{Design principles.}
The student is designed for predictable inference latency and memory usage in on-premises environments. We preserve the overall nnU-Net architecture (resolution hierarchy and skip connections) to maintain compatibility with the same preprocessing pipeline and deployment interface used by the teacher. This continuity reduces engineering overhead during integration and simplifies validation when swapping model variants.

\paragraph{Capacity scaling under clinical budgets.}
Instead of using a single fixed student size, we define a small family of students by uniformly scaling the channel width of the nnU-Net backbone. Let $\alpha \in \{1/2, 1/4\}$ denote the channel-width scaling factor, where $\alpha=1/2$ and $\alpha=1/4$ correspond to half-width and quarter-width students. All convolutional layers in the student use $\alpha$ times the original channel width of the teacher configuration.

Because parameters and FLOPs in convolutional networks scale approximately quadratically with channel width, this uniform scaling yields an approximate $\mathcal{O}(\alpha^2)$ reduction in model capacity while preserving the computation graph and feature hierarchy. This property is important for maintainability: different student capacities can be deployed through the same runtime environment and workflow interface without additional system changes.

\paragraph{Impact on deployment.}
The student family enables hardware-aligned deployment choices. For example, $\alpha=1/4$ prioritizes throughput and lower memory usage, while $\alpha=1/2$ offers a stronger accuracy-efficiency trade-off. In all experiments, we compare distilled and non-distilled students at the same $\alpha$ to isolate the benefit of distillation from architectural capacity.

\subsection{Logit-based Knowledge Distillation Framework}

We adopt a logit-based distillation objective that aligns the student and teacher output distributions using a temperature $\tau$. For each voxel, the softened class probabilities are
\begin{equation}
\begin{aligned}
\mathbf{p}_T &= \mathrm{softmax}(\mathbf{z}_T/\tau), \\
\mathbf{p}_S &= \mathrm{softmax}(\mathbf{z}_S/\tau).
\end{aligned}
\end{equation}

The distillation loss is defined as the voxel-wise Kullback--Leibler (KL) divergence
between the teacher and student probability distributions:
\begin{equation}
\mathcal{L}_{\mathrm{KD}} =
\tau^2 \frac{1}{|\Omega|}
\sum_{u \in \Omega}
\sum_{c=1}^{C}
p_T^{(u,c)} \log
\frac{p_T^{(u,c)}}{p_S^{(u,c)}},
\end{equation}

where $\Omega$ denotes the set of voxels and $C$ the number of classes.
This loss encourages the student to match the teacher’s softened class
distribution at each voxel, capturing inter-class relationships that are
not available from hard labels alone.

The student is trained with a combined objective:
\begin{equation}
\mathcal{L}_{\mathrm{total}} = \mathcal{L}_{\mathrm{seg}} + \lambda \, \mathcal{L}_{\mathrm{KD}},
\end{equation}
where $\mathcal{L}_{\mathrm{seg}}$ is the standard supervised segmentation loss (Dice + cross-entropy) in nn-Unet, and $\lambda$ controls the contribution of distillation.

\subsection{Evaluation Metrics}

Model performance is evaluated using the Dice similarity coefficient (Dice), normalized surface Dice (NSD), and the 95th percentile Hausdorff distance (HD95). Dice quantifies volumetric overlap between predicted and reference segmentations and is commonly used to assess overall segmentation accuracy. In contrast, NSD and HD95 are boundary-sensitive metrics that measure surface agreement and worst-case surface deviation, respectively.

These complementary metrics provide a balanced evaluation of both volumetric fidelity and boundary accuracy, which is particularly important in clinical segmentation tasks where small structural errors or boundary inconsistencies may disproportionately affect downstream measurements, visualization, and clinical interpretation.

\subsection{Deployment Pathway within Clinical Workflows}

After training, the student model is packaged for inference as a lightweight service that can be executed on an on-premises inference node. The deployment pathway is designed to fit standard clinical workflows: imaging studies are retrieved through PACS, processed locally, and the segmentation outputs are returned to downstream systems for visualization or quantitative analysis. Since the student preserves the teacher’s preprocessing and I/O interface, model updates can be performed by re-distillation and replacement of the student checkpoint without changing the surrounding workflow components.

\section{Data and Code availability}
The BTCV dataset used in this study is publicly available from Synapse at \url{https://www.synapse.org/#!Synapse:syn3193805/wiki/}.
Other brain-related datasets are subject to copyright and licensing restrictions and are therefore not publicly available, but may be obtained from the data owners upon reasonable request.
\section{Code availability}
The source code for distillation training and evaluation is available at \url{https://github.com/lanqz7766/nnUNet-KD}.

\section*{Acknowledgements}

XJ is a CPRIT Scholar in Cancer Research (RR180012) and was supported in part by the Christopher Sarofim Family Professorship, the UT STARS Award, UTHealth Houston startup funds, and grants from the National Institutes of Health (NIH), including R01AG066749, R01AG066749-03S1, R01LM013712, R01LM014520, R01AG082721, U01AG079847, U24LM013755, U01CA274576, and U54HG012510. This work was also supported in part by the National Science Foundation (NSF) under grant number 2124789.

\bibliography{sn-bibliography}

\end{document}